\documentclass[acmsmall,screen,natbib=False]{acmart}

\usepackage{adjustbox}
\usepackage{threeparttable}
\usepackage{booktabs}
\usepackage{algorithm}
\usepackage{algorithmic}
\usepackage{tabularx}
\usepackage{amsmath,amsfonts}
\usepackage{stfloats}
\usepackage{makecell} 
\usepackage{pgfplots}
\pgfplotsset{compat=1.17}
\usepackage[unicode]{hyperref}
\usepackage{xcolor, colortbl}
\usepackage{soul}
\usepackage{placeins}
\RequirePackage[
  datamodel=acmdatamodel,
  style=acmnumeric, 
]{biblatex}

\AtBeginDocument{%
  }

\acmJournal{TOMM}

\begin{document}
\title{CapST: Leveraging Capsule Networks and Temporal Attention for Accurate Model Attribution in Deep-fake Videos}

\author{Wasim Ahmad}
\email{was_last@iis.sinica.edu.tw}
\orcid{0000-0002-4841-6481}
\affiliation{
  \institution{Institute of Information Science, Academia Sinica}
  \city{Taipei}
  \country{Taiwan (R.O.C.)}
}
\affiliation{
  \institution{Social Networks and Human-Centred Computing, Taiwan International Graduate Program}
  \city{Taipei}
  \country{Taiwan (R.O.C.)}
}
\affiliation{
  \institution{Department of Computer Science, National Chengchi University}
  \city{Taipei}
  \country{Taiwan (R.O.C.)}
}

\author{Yan-Tsung Peng}
\email{ytpeng@cs.nccu.edu.tw}
\affiliation{
  \institution{\textit{Member IEEE}, National Chengchi University}
  \city{Taipei}
  \country{Taiwan (R.O.C)}
}

\author{Yuan-Hao Chang}\authornote{Corresponding Author: Yuan-Hao Chang}
\email{johnson@iis.sinica.edu.tw}
\affiliation{
  \institution{\textit{Fellow IEEE}, Institute of Information Science, Academia Sinica}
  \city{Taipei}
  \country{Taiwan (R.O.C)}
}

\author{Gaddisa Olani Ganfure}
\email{gaddisaolex@gmail.com}
\affiliation{%
  \institution{Dire Dawa University}
  \city{Dire Dawa}
  \country{Ethiopia}
}

\author{Sarwar Khan}
\email{sarwar@iis.sinica.edu.tw}
\affiliation{
  \institution{Research Center for Information Technology Innovation, Academia Sinica}
    \city{Taipei}
  \country{Taiwan (R.O.C.)}
}
\affiliation{
  \institution{Social Networks and Human-Centred Computing, Taiwan International Graduate Program}
  \city{Taipei}
  \country{Taiwan (R.O.C.)}
}
\affiliation{
  \institution{Department of Computer Science, National Chengchi University}
  \city{Taipei}
  \country{Taiwan (R.O.C.)}
}

\renewcommand{\shortauthors}{W.Ahmad et al.}

\begin{abstract}
Deep-fake videos, generated through AI face-swapping techniques, have garnered considerable attention due to their potential for impactful impersonation attacks. While existing research primarily distinguishes real from fake videos, attributing a deep-fake to its specific generation model or encoder is crucial for forensic investigation, enabling precise source tracing and tailored countermeasures. This approach not only enhances detection accuracy by leveraging unique model-specific artifacts but also provides insights essential for developing proactive defenses against evolving deep-fake techniques. Addressing this gap, this paper investigates the model attribution problem for deep-fake videos using two datasets: Deepfakes from Different Models (DFDM) and GANGen-Detection, which comprise deep-fake videos and images generated by GAN models. We select only fake images from the GANGen-Detection dataset to align with the DFDM dataset, which specifies the goal of this study, focusing on model attribution rather than real/fake classification. This study formulates deep-fake model attribution as a multiclass classification task, introducing a novel Capsule-Spatial-Temporal (CapST) model that effectively integrates a modified VGG19 (utilizing only the first 26 out of 52 layers) for feature extraction, combined with Capsule Networks and a Spatio-Temporal attention mechanism. The Capsule module captures intricate feature hierarchies, enabling robust identification of deep-fake attributes, while a video-level fusion technique leverages temporal attention mechanisms to process concatenated feature vectors and capture temporal dependencies in deep-fake videos. By aggregating insights across frames, our model achieves a comprehensive understanding of video content, resulting in more precise predictions. Experimental results on the DFDM and GANGen-Detection datasets demonstrate the efficacy of CapST, achieving substantial improvements in accurately categorizing deep-fake videos over baseline models, all while demanding fewer computational resources.
\end{abstract}

\setcopyright{acmlicensed}
\acmJournal{TOMM}
\acmYear{2025} \acmVolume{1} \acmNumber{1} \acmArticle{1} \acmMonth{1}\acmDOI{10.1145/3715138}



\received{2024}
\received[Revised]{2024}
\received[Accepted]{2024}
\received[Published]{2025}

\maketitle


\section{INTRODUCTION}
\begin{figure}[htbp]
    \centering
     \begin{adjustbox}{trim={1.8cm} {5.5cm} {1.8cm} {4.5cm},clip, width=\linewidth}
    \includegraphics{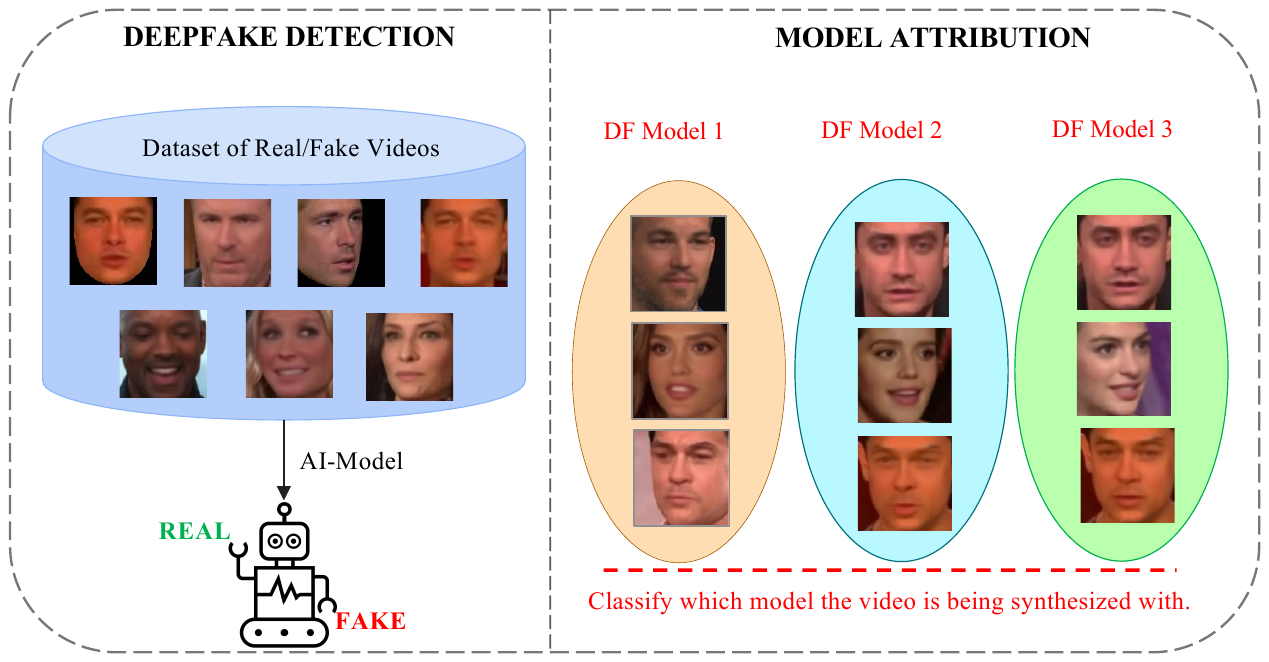}
  \end{adjustbox}
  \Description{In contrast to prevailing methods designed to differentiate between authentic and manipulated videos, deep-fake model attribution seeks to discern the specific generation model employed in a given deep-fake video. This process is instrumental in tracing the video's authorship and source.}
    \caption{Model Attribution vs Model Detection}
    \label{fig_model_attrib}
\end{figure}

The term ``Deepfake'' derived from ``deep learning'' and ``fake'' commonly denotes AI-generated face-swap videos, wherein the faces of the original subject (the source) are replaced with those of another person (the target). Generated using deep learning models with the same facial expressions as the source, these videos have become increasingly easy to produce with the advent of open source generation tools \cite{9105991}. The potential weaponization of Deepfakes by malicious actors to manipulate identities and spread disinformation has raised widespread concerns [2]. Such malicious videos are forged via facial manipulation techniques, commonly termed deep-fake videos, and have emerged as substantial threats to privacy and security. These deceptive videos are effortlessly produced using accessible software like Deepfakes \cite{faceswap} and DeepFaceLab \cite{deepFaceLab}. Hence, creating a powerful and resilient deep-fake detection methodology becomes paramount. Interestingly, a multitude of deep-fake detectors, including \cite{li2020face,gu2021spatiotemporal,yu2022improving} have been formulated and demonstrated effectiveness in expansive deep-fake datasets \cite{rossler2019faceforensics++} \cite{li2020celeb} \cite{dolhansky2019deepfake}. 
In response, there has been a growing research interest in deep-fake detection methods, especially current face forgery technologies, which encompass various techniques, including non-target face generation \cite{park2020contrastive,karras2020analyzing}, expression replay \cite{burkov2020neural,song2021pareidolia}, face replacement \cite{agarwal2021detecting,zhu2021one} and attribute editing \cite{fu2021high,xu2021facecontroller}. Existing techniques often leverage visual / signal artifacts \cite{guarnera2020deepfake,mittal2020emotions} or novel deep neural networks \cite{kim2021fretal,8682602,de2020deepfake,afchar2018mesonet}. Notable datasets of deep-fake videos have been curated, including Faceforensics++ \cite{rossler2019faceforensics++}, Deeperforensics-1.0 \cite{jiang2020deeperforensics}, DFDC \cite{dolhansky2019deepfake}, WildDeep-fake \cite{zi2020wilddeepfake} and Celeb-DF \cite{li2020celeb}. Although state-of-the-art detection methods have shown promising performance in benchmark datasets, the specific methods employed in the creation of deep-fake are crucial for forensic analysis, as shown in Figure \ref{fig_model_attrib}. Knowing the synthesis model used can help narrow down potential users who have downloaded publicly available tools.
\begin{figure}[htbp]
  \centering
  \begin{adjustbox}{trim={4cm} {5cm} {4cm} {4.7cm},clip, width=\linewidth}
    \includegraphics{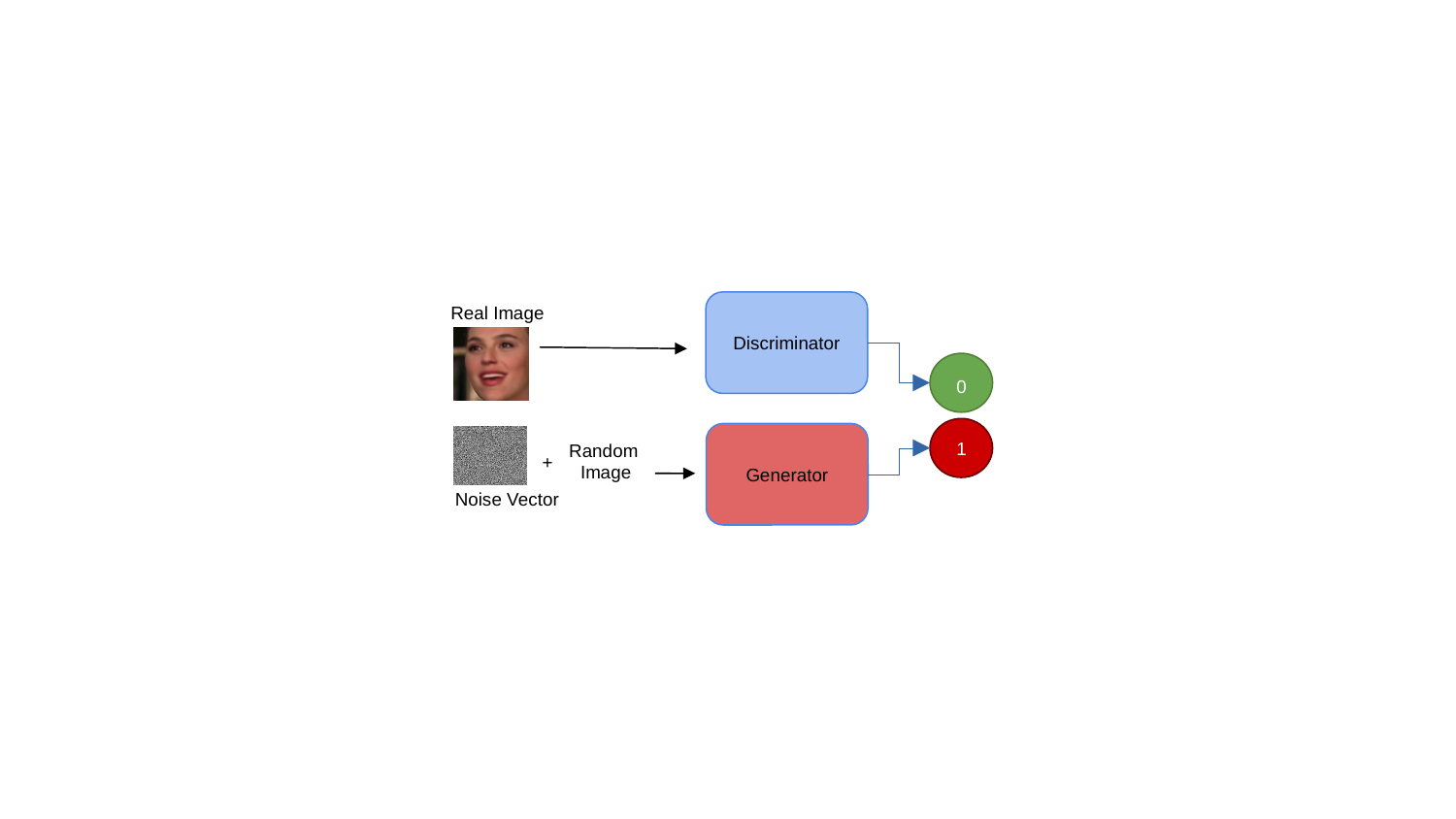}
  \end{adjustbox}
  \caption{Generative Adversarial Networks (GANs}
  \Description{The process of Generative Adversarial Networks (GANs) to produce lifelike and top-notch data involves training two neural networks, specifically a generator and a discriminator, within an adversarial framework.}
  \label{gans}
\end{figure}
\begin{table}[!b]
  \centering
  \caption{Structure of DFDM dataset \cite{jia2022model}}
  \label{tab_dfdm_data_generation}
  \begin{tabularx}{\textwidth}{lcccc>{\raggedright\arraybackslash}X}
  \hline
    Model & Input & Output & Encoder & Decoder & Variation \\
    \hline
    Faceswap    & 64 & 64 & 4Conv+1Ups & 3Ups+1Conv & / \\
    Lightweight & 64 & 64 & 3Conv+1Ups & 3Ups+1Conv & Encoder \\
    IAE         & 64 & 64 & 4Conv & 4Ups+1Conv & Intermediate layers Shared Encoder Decoder \\
    Dfaker      & 64 & 128 & 4Conv+1Ups & 3Ups+3Residual+1Conv & Decoder \\
    DFL-H128    & 128 & 128 & 4Conv+1Ups & 3Ups+1Conv & Input Resolution \\
    \hline
  \end{tabularx}
\end{table}

\noindent The model attribution problem, crucial to understanding the Deepfakes creation process, has been explored in recent studies concerning models based on Generative Adversarial Network (GAN) \cite{girish2021towards,yu2019attributing}, as shown in Figure \ref{gans}. However, the model-attribution methods developed for GAN images are not directly applicable to face-swap deep-fake videos. This distinction presents a more challenging scenario, particularly because autoencoder-based models, commonly used in deep-fake creation \cite{dolhansky2020deepfake}, attenuate high-frequency features in generated video frames, a characteristic on which previous methods for GAN model attribution are based. DFDM is generated using various autoencoder models, as illustrated in Table \ref{tab_dfdm_data_generation}. DFDM comprises five balanced categories of deep-fake videos created through Facewap \cite{faceswap}, with variations in encoder, decoder, intermediate layer, and input resolution. Similarly, GANGen-Detection, a dataset specifically generated for generalizable deep-fake detection tasks, contains a variety of fake images generated by GAN on multiple GAN architectures, providing a wider set of synthetic manipulation techniques for evaluation \cite{aigcdataset}. Together, these datasets allow us to explore the attribution of the fake model in different generation techniques, using only fake data to focus exclusively on the model attribution objective.
\begin{figure}[htbp]
    \centering
     \begin{adjustbox}{trim={1cm} {4cm} {1cm} {4.5cm},clip, width=\linewidth}
    \includegraphics{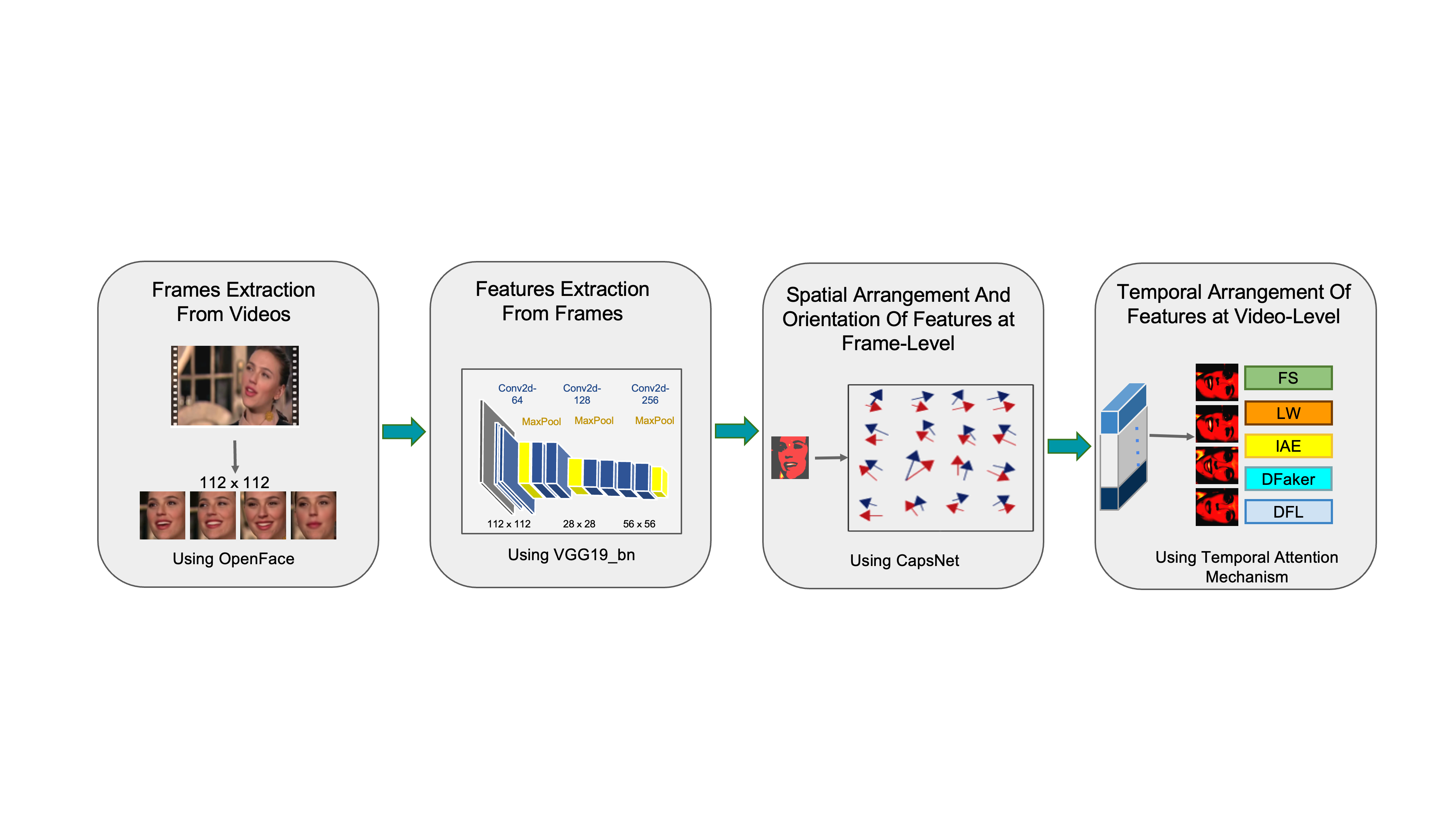}
  \end{adjustbox}
    \caption{General Overview of CapST}
    \Description{First, we extract the distinctive attributes of a set of video frames using VGG19. Subsequently, these resultant features are fed into the capsule module to delineate spatial configurations within each frame. At last, these refined features are directed into the video-level fusion module, thereby enabling the classification of the manipulated class to which the video pertains.}
    \label{general_overview}
\end{figure}
In this study, we propose an innovative approach to address the challenge of attribution of the deep-fake model by introducing CapST, an enhanced and lightweight model that combines the strengths of Capsule Networks \cite{sabour2017dynamic} and a modified spatial temporal attention mechanism. In particular, this is only the second attempt in this specific domain, based on the previous work presented in \cite{jia2022model} and illustrated in Figure \ref{general_overview}. Our key contributions are as follows.

\begin{itemize} 
    \item We present the Capsule Spatial-Temporal (CapST) model for deep-fake model attribution. CapST leverages the initial three layers of VGG19 for efficient feature extraction, incorporates Capsule Networks as a self-attention module in both spatial and temporal dimensions, and introduces a novel self-subtract mechanism to enhance the detection of temporal artifacts.
    
    \item Our model demonstrates versatility beyond model attribution, with potential applicability across a range of classification tasks (also not addressed here due to the domain-specific study).
    
    \item Designed with resource efficiency in mind, the CapST model avoids dense configurations, ensuring lower computational costs, a critical consideration when training large models.
    
    \item Experimental results confirm the effectiveness of CapST in addressing model attribution, achieving significant performance improvements while maintaining a resource-efficient profile, with only 3.27 million parameters required for operation.
\end{itemize}

\section{BACKGROUND AND RELATED WORK}
\subsection{Video/Image Forgery Detection}
In recent years, there has been growing interest in the attribution and detection of AI-generated content, particularly with the rise of sophisticated generative models. For instance, \cite{sha2023fake} explores the detection and attribution of fake images created by text-to-image models, a burgeoning area of research in AI-generated content attribution. Similar to our work on deep-fake video attribution, this study highlights the importance of identifying model-specific features in generative models, but it focuses on the attribution of static images rather than the temporal complexities present in video forgeries. Similarly, \cite{ojha2023towards} addresses the generalization of detectors across various generative models, emphasizing the limitations of current classifiers trained on one model when tested on others (e.g., GANs vs. diffusion models). While their focus is on generalizing detection methods across static content, however, our study contributes to this field by tackling the complexities in video-based deepfakes, where temporal coherence across frames adds another layer of difficulty for model attribution. \cite{keita2025fidavl} in contrast, focused on the detection and attribution of fake images using a multimodal approach, combining visual and textual data. This aligns with our work's focus on attribution but differs in its focus on static images. Our study goes a step further by addressing the unique challenges of deep-fake videos, specifically handling temporal inconsistencies across frames, something that static image-based methods like FIDAVL \cite{keita2025fidavl} do not cover. In the early stages of face generation technologies, direct visual artifacts and inconsistencies in facial areas were common due to their uncontrollable characteristics. Image forgery detection efforts often leveraged statistical inconsistencies in faces or heads. For instance, \cite{li2020face} proposed Face X-ray, a method predicting blending boundaries in fake video frames. \cite{zhang2022local} employed adversarial learning to add perturbations to regions of interest, revealing vulnerabilities in current image-based forgery detection. \cite{yang2021msta} tracked potential texture traces during image generation, performing well on multiple benchmarks. \cite{li2022artifacts} introduced an Artifacts-Disentangled Adversarial Learning framework for accurate deep-fake detection. Leveraging the Multi-scale Feature Separator (MFS) and Artifacts Cycle Consistency Loss (ACCL). \cite{chen2022learning} demonstrated good generalization abilities through intra-consistency within classes and inter-diversity between classes. However, with the advancement of deep-fake generation techniques, image-based detectors may fail to capture temporal inconsistencies across multiple frames, as early forgery generation technology showed obvious temporal inconsistency between frames. \cite{hu2021detecting} also introduced a spatial and temporal level two-stream detection network, significantly improving post-processing compressed low-pixel videos. \cite{zheng2021exploring} proposed a temporal convolution network effective in extracting temporal features and improving generalization. \cite{ganiyusufoglu2020spatio} developed a forgery detection method capturing local spatiotemporal relations and inconsistencies in deep-fake videos. \cite{wang2021attention} proposed AGLNet, aiming to improve deep-fake detection by considering spatial and temporal correlations. AGLNet employs attention-guided LSTM modules to aggregate detailed information from consecutive video frames. \cite{das2021demystifying} explore and adapt attention mechanisms for deep-fake detection, specifically focusing on learning discriminative features that consider the temporal evolution of faces to identify manipulations. The research addresses the questions of how to use attention mechanisms and what type of attention is effective for deep-fake detection. \cite{chen2022deepfake}  introduces a neural deep-fake detector aiming to identify manipulative signatures in forged videos at both frame and sequence levels. Using a ResNet backbone with spatial and distance attention mechanisms, the model is trained as a classifier to detect forged content. \cite{yu2023msvt} Introduced MSVT, a novel Multiple Spatiotemporal Views Transformer with Local (LSV) and Global (GSV) views, to extract detailed spatiotemporal information. Utilizing local-consecutive temporal views for LSV and a global-local transformer (GLT), MSVT effectively integrates multi-level features for comprehensive information mining. \cite{yin2023dynamic} proposed a dynamic difference learning method for precise spatio-temporal inconsistency in deep-fake detection, employing a dynamic fine-grained difference capture module (DFDC-module) with self-attention and denoising, along with a multi-scale spatio-temporal aggregation module (MSA-module) for comprehensive information modeling. This approach extends 2D CNNs into dynamic spatio-temporal inconsistency capture networks. 

\begin{table}[!t]
  \centering
  \caption{Popular DF Models, Datasets and its parameters}
  \label{popular_df_dataset_model}
  \begin{tabular}{lcc}
    \hline
    \textbf{Model} & \textbf{Datasets} & \textbf{Params(M)}\\
    \hline
    X-Ray \cite{li2020face}    & FF++ $|$ UADFV $|$CDF\_v1,v2 & 20.8\\
    TSN \cite{zheng2021exploring}    & FF++ $|$ UADFV $|$CDF\_v1,v2 & 26.6\\
    Two-stream \cite{zhou2017two} & FF++ $|$ UADFV $|$CDF\_v1,v2 & 23.9\\
    MesoInception4 \cite{afchar2018mesonet} & FF++ $|$ UADFV $|$CDF\_v1,v2 & 28.6\\
    FWA \cite{li2019exposing} & FF++ $|$ UADFV $|$CDF\_v1,v2 & 25.6\\
    Xception-raw,c-23,c40 \cite{rossler2019faceforensics++} & FF++ $|$ UADFV $|$CDF\_v1,v2 & 22.9\\
    Capsule \cite{8682602} & FF++ $|$ UADFV $|$CDF\_v1,v2 & 3.9\\
    Multi-attentional \cite{zhao2021multi} & FF++ $|$ UADFV $|$CDF\_v1,v2 & 19.5\\
    FTCN \cite{zheng2021exploring}    & FF++ $|$ UADFV $|$CDF\_v1,v2 & 26.6\\
    RealForensics \cite{tran2019video}    & FF++ $|$ UADFV $|$CDF\_v1,v2 & 21.4\\
    ResVit \cite{ahmad2022resvit}    & FF++ $|$   DFD   $|$CDF\_v1,v2 & 21.4\\
    Cross Eff\_ViT \cite{coccomini2022combining}          & DFDC  & 109\\
    Selim EffNet B7 \cite{seferbekov2020dfdc} & DFDC & 462\\
    Convolutional ViT \cite{wodajo2021deepfake}   & DFDC & 89\\
    Multimodal \cite{9980255}    & FakeAVCeleb &  33.6\\ 
    \hline
  \end{tabular}
\end{table}

\subsection{Popular Deep-fake models and Datasets}
Recent strides in fake technology result from the advancement of sophisticated models and the availability of well-curated datasets. These instrumental components have significantly advanced the frontiers of deep-fake synthesis, detection, and analysis. Earlier discussions highlighted numerous studies presenting outstanding AI models that exhibited superior performance across diverse datasets, as detailed in the subsequent section. However, a prevalent challenge among these models lies in their size and complexity, leading to heightened daily resource demands. The following section provides insights into the famous deep-fake detection models and datasets, which include their resource consumption parameters. Refer to Table \ref{popular_df_dataset_model} for specific examples of these models and datasets.

\subsubsection{Deep-fake Models}
XceptionNet \cite{rossler2019faceforensics++} is a deep CNN architecture with high performance in deep-fake detection, using separable convolutions in depth for fine-grained feature capture. DeepFakeTIMIT \cite{bekci2020cross} specializes in audio-visual deep-fake synthesis, combining audio and visual information. Ensemble Models combine predictions from diverse models, achieving robust deep-Fake detection. SqueezeNAS \cite{ding2021hr} is an automated design framework for compact models optimized for accuracy and efficiency. FWA-Net focuses on detecting facial warping artifacts using a two-stream architecture. Capsule-Forensics \cite{8682602} uses capsule networks for forged image and video detection. MesoNet \cite{afchar2018mesonet} is a lightweight CNN for the detection of mesoscopic levels. DFDNet \cite{li2020blind} employs a two-stream architecture for spatiotemporal features.

\subsubsection{Datasets}
A variety of deep-fake datasets are currently available, each serving different aspects of research in deep-fake detection and synthesis. Some of the most notable datasets include:

\begin{itemize}
    \item \textbf{FaceForensics++} \cite{rossler2019faceforensics++}: A benchmark dataset containing over 1,000 manipulated facial videos, widely used for deep-fake detection research.
    \item \textbf{CelebA} \cite{liu2015deep}: A large-scale celebrity image dataset commonly used in facial attribute recognition and deep-fake studies.
    \item \textbf{DFDC Dataset} \cite{dolhansky2006deepfake}: Created for the deep-fake Detection Challenge, this dataset focuses on state-of-the-art deep-fake video detection.
    \item \textbf{Face2Face} \cite{thies2016face2face}: Specializes in manipulating facial expressions in videos, facilitating research in real-time face reenactment.
    \item \textbf{VoxCeleb Dataset} \cite{nagrani2017voxceleb}: Provides a diverse collection of audio-visual celebrity speech data, useful for cross-modal learning and deep-fake synthesis.
    \item \textbf{DeepfakeTIMIT} \cite{bekci2020cross}: Designed for audio-visual synthesis, this dataset includes video sequences paired with corresponding audio.
    \item \textbf{FBDB} \cite{dolhansky2019deepfake}: Originally released by Facebook, this large-scale dataset supports training and evaluation of deep-fake detection.
\end{itemize}

\noindent Although these data sets are instrumental in addressing deep-fake detection, our research focuses on model attribution, for which the \textbf{DFDM} dataset \cite{jia2022model} is particularly suitable. The DFDM dataset is specifically designed to facilitate research in identifying the source models used to generate deep-fake content.

\subsection{Model Attribution}
Model attribution involves the crucial task of identifying the specific model used to generate synthetic images or videos \cite{girish2021towards}. Within the domain of image manipulation, the widespread use of GAN models has spurred numerous investigations \cite{yu2019attributing, marra2019gans} into the attribution of GAN models, particularly in the field of counterfeit images. These attribution methodologies aim to extract distinctive artificial markers left by diverse GAN models, employing multiclass classification techniques to attribute the specific GAN model used. However, until the work by \cite{jia2022model}. Prior research had not concentrated on model attribution specifically for deep-fake face-swap videos. Although GAN models are capable of producing Deepfakes, their effectiveness is often restricted to controlled settings with uniform lighting conditions \cite{dolhansky2019deepfake}. Consequently, the generation of publicly available face-swap Deepfakes frequently relies on deep-Fake Autoencoder (DFAE) techniques. The feasibility of differentiating between various DFAE models for deep-fake attribution and the extent to which image-based GAN model attribution techniques apply to DFAE models in the video domain remain subjects warranting exploration. These inquiries are pivotal in advancing deep-fake model attribution, addressing research gaps, and enhancing our understanding of challenges in attributing models for face-swapping techniques. The exploration of attribution methodologies across generative models and their application to various video generation techniques contributes to ongoing efforts to improve the reliability and robustness of model attribution in the evolving landscape of synthetic media \cite{girish2021towards, yu2019attributing, marra2019gans, jia2022model, dolhansky2019deepfake}.


\section{CapST ARCHITECTURE}
This study emphasizes the design of a resource-efficient Deep Neural Network (DNN) for targeted domains. Our CapST Model, detailed in this section, combines Capsule Network with Spatial and Temporal attention mechanisms for deep-fake content analysis. Using frames extracted, using VGG\_19 for feature extraction, and employing a capsule network with MFM activation and spatial attention, our algorithm enhances artifact detection. TAM ensures adaptive feature aggregation across frames, enhancing accuracy in deep-fake classification (See Algorithm \ref{capst_algorithm}).

\begin{figure*}
  \centering
  \begin{adjustbox}{trim={0cm} {2cm} {0cm} {4cm},clip, width=\linewidth}
    \includegraphics{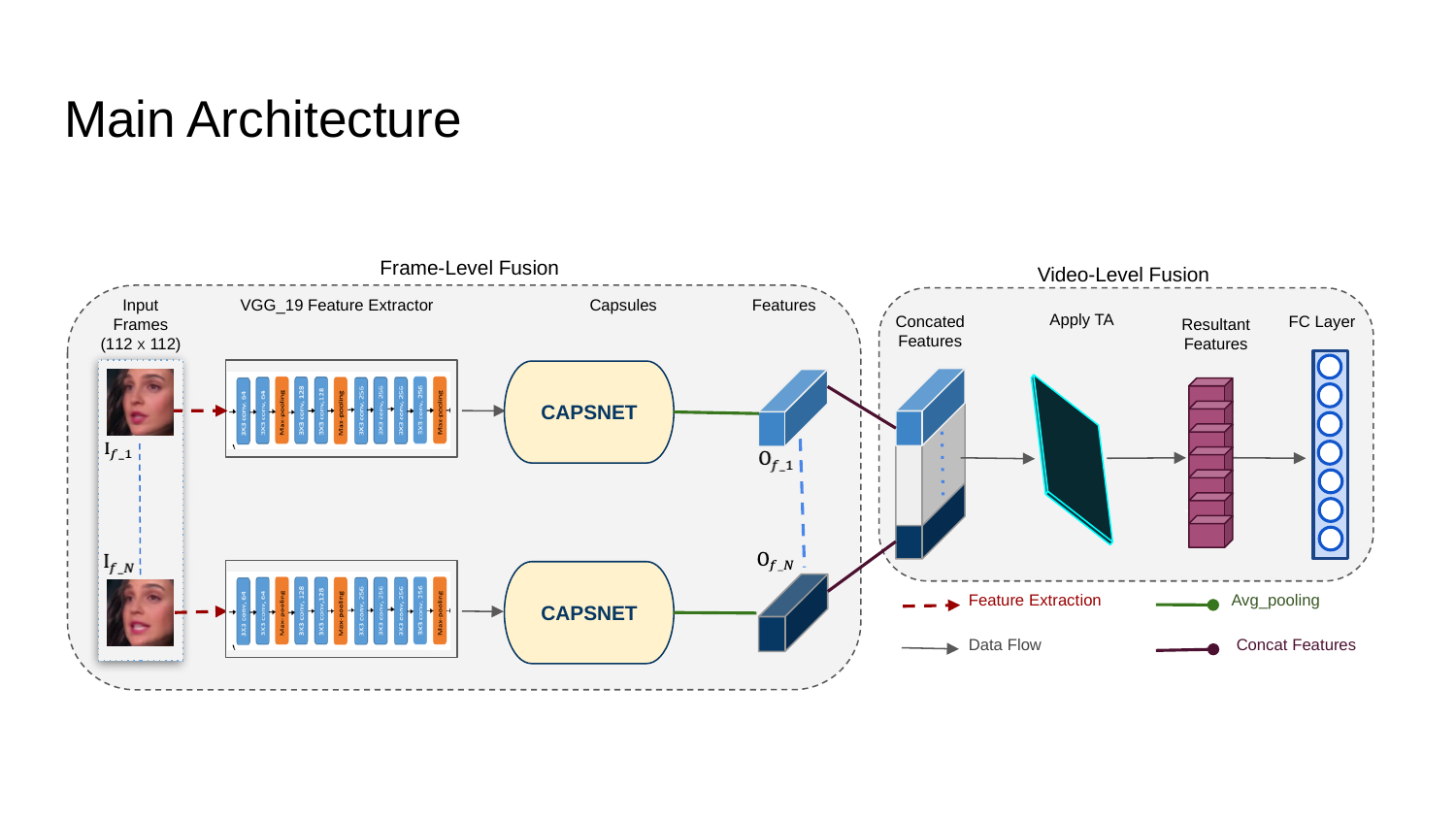}
  \end{adjustbox}
  \caption{CapST Main architecture}
  \Description{The model leverages the VGG19 model as a backbone to extract facial features from video frames and incorporates a Capsule Network (CapsuleNet) with dynamic routing to capture temporal dependencies among these features over time. Additionally, it utilizes a Temporal Attention Map (TAM) to enhance the model's ability to focus on relevant information across multiple frames, providing a comprehensive approach for deep-fake video model attribution by considering both spatial and temporal aspects in the feature representation.}
  \label{fig_4}
\end{figure*}
\noindent
\subsection{Unveiling the intricacies of DF artifacts at frame-level}
\begin{figure*}
  \centering
  \begin{adjustbox}{trim={0cm} {2.3cm} {0cm} {5cm},clip, width=\linewidth}
    \includegraphics{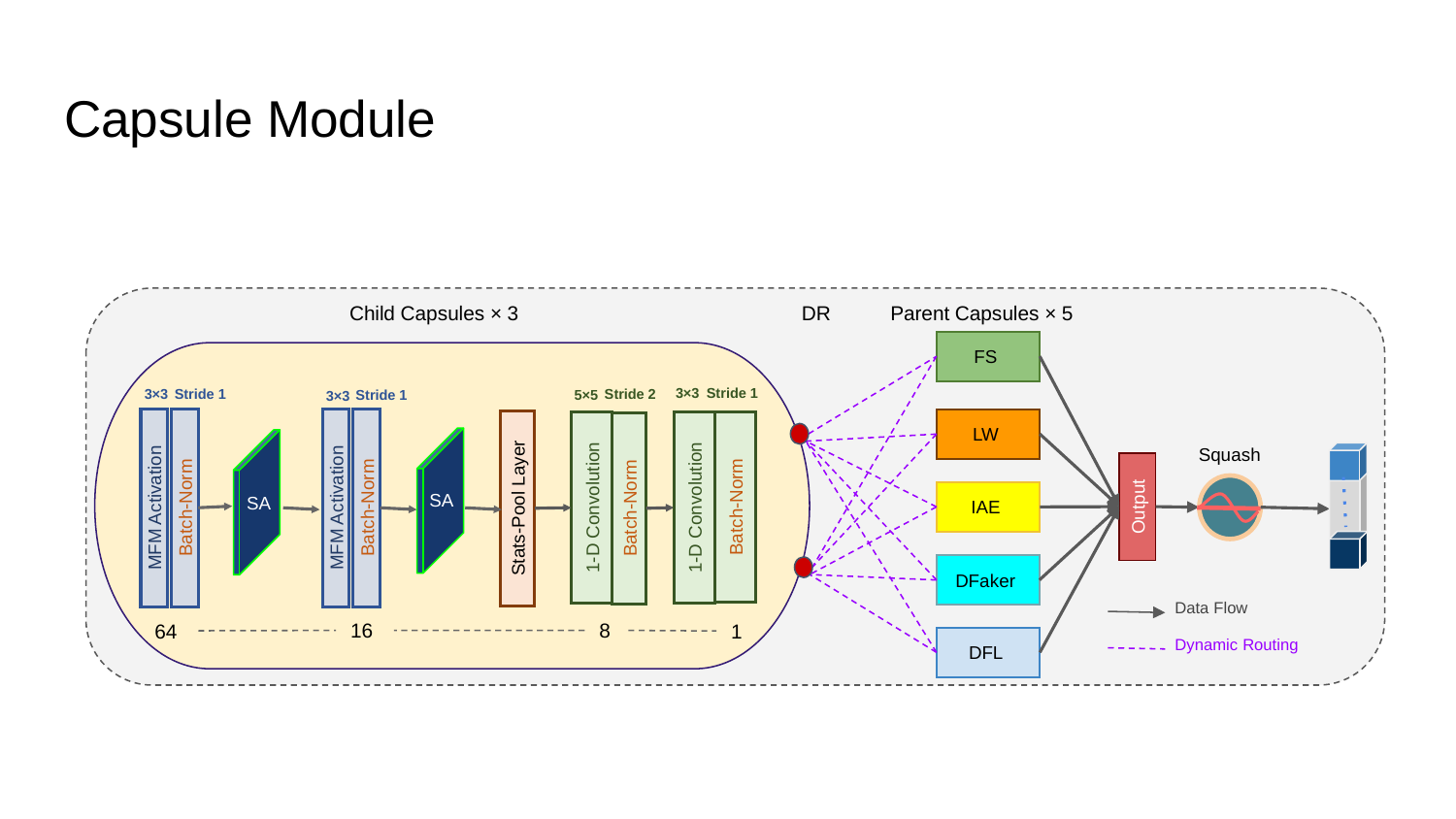}
  \end{adjustbox}
  \caption{Capsule Module of the CapST Architecture}
  \Description{The capsule module starts by applying the MFM activation function for lightweight and robust network behavior, suppressing low activation neurons. Subsequently, a spatial attention layer aggregates frame features and a Statistical pooling layer calculates the mean and variance for each filter/kernel. The resulting data is then processed through a 1D convolutional segment, followed by dynamic routing to output capsules for multi-class classification. The ultimate output is determined based on the activation patterns of the output capsules, contributing to the comprehensive functioning of the capsule module in deep-fake video model attribution.}
  \label{fig_5}
\end{figure*}

The architecture of our CapST model is depicted in Figure \ref{fig_4}. We first extract N number of faces from each video in the dataset, i.e., 
\begin{equation}
    V_i =I_{f1}, I_{f2}, \ldots, I_{fN}, \label{eq:inputframes}
\end{equation}
where $V_i$ represents the input video while $I_{f1}, \ldots, I_{fN}$ represents the number of input frames we chose to consider for each video. After that, we pass these frames one by one to the VGG\_19 network, where we extract the features of the input frame. We choose VGG\_19 as a feature extractor pre-trained on \cite{russakovsky2015imagenet} from the first-to-third Max-Pooling layer with a total of 26 layers. VGG\_19 is used as its performance is better throughout all the experiments than other CNN networks in this specific task.  The extracted features are then passed through the Capsule Network as shown in Figure \ref{fig_5}. 
We divided each primary capsule into three sections, including the Max Feature Map (MFM)~\cite{wu2018light} activation function with the combination of Batch Normalization and Spatial Attention Layer (SA), statistical grouping layer and a one-dimensional CNN. The statistical pooling layer is considered effective for forensic tasks and can improve the performance of the network \cite{rahmouni2017distinguishing,nguyen2018modular} by making the network independent even if the dimensions of the input images vary. This design is based on methods that have been shown to work well for flexible, task-independent models, which means the network could potentially be used for different tasks without being completely redesigned. However, testing on multiple tasks is beyond the scope of this paper. 
The capsule module first passes the input features through the MFM activation function proposed in \cite{wu2018light}, which enables the network to become light and robust as it suppresses low-activation neurons in each primary capsule. The MFM output is then passed through a spatial attention layer to aggregate the frame features \cite{woo2018cbam}. The statistical pooling layer is then used to calculate the mean and variance of each filter/kernel, as shown below. 
\parskip 0.1in
\begin{itemize}
\setlength\itemsep{1em}
\item Mean: 
\begin{equation} \mu_k = (1 / H \times W - 1)\sum_{i=1}^{H}\sum_{j=1}^{W}I_kij \label{eq:mean} \end{equation}
\item Variance: \begin{equation}
\sigma^2_k = (1 / H \times W - 1)\sum_{i=1}^{H}\sum_{j=1}^{W}(I_kij - \mu_k)^2 \label{eq:variance}
\end{equation}  
\end{itemize}

\noindent Here, the variable \( k \) denotes the index of the layer, while \( H \) and \( W \) represent the height and width of the filter, respectively. \( I \) refers to a two-dimensional array of filters. The outcome of the statistical layer lends itself well to 1D convolution. After traversing the ensuing 1D convolutional segment, the data undergo dynamic routing to reach the output capsules. The ultimate outcome is determined based on the activation status of the output capsules. This process is followed by the presence of five output capsules designated for multiclass classification, depicted in Fig. \ref{fig_5}. The ultimate output is derived from the activation patterns exhibited by the output capsules.

\subsection{Video-level DF Artifact Analysis}
The output of each frame, as shown in Equation \ref{eq:inputframes}, is combined to represent a video instance, as illustrated in Equation \ref{eq:outputframes}.
\begin{equation}
    V_o = O_{f1}, O_{f2}, \ldots, O_{fN}, \label{eq:outputframes}
\end{equation}
where $V_o$ is the output video, and $O_{f1}, \ldots, O_{fN}$ are the output frames from 1 to $N$. Previous studies mainly used score fusion for deep-fake detection at video level \cite{li2019exposing, afchar2018mesonet, 8682602, jia2022model, li2021frequency}. Our method enhances classification by using a temporal attention map (TAM) for adaptive feature aggregation. We concatenate frame features to form multi-frame representations, which are fused using a TAM similar to SENet \cite{hu2018squeeze}. TAM employs ReLU activation and two fully connected layers, and its output is fed into another fully connected layer for classification, producing class probabilities for the five deep-fake categories. The network is trained with a cross-entropy loss function:
\vspace{0cm}
\begin{equation}
L(\mathbf{y}, \mathbf{\hat{y}}) = - \sum_{i=1}^{5} y_i \cdot \log(\hat{y}_i) \label{eq:lossfunction}
\end{equation}
\vspace{0cm}
\noindent Here, \( y \) is the true label vector and \( \hat{y} \) is the predicted probability vector for five classes. This approach leverages the Capsule Network's attention mechanism to extract subtle frame-level variations for model attribution. Unlike GAN model attribution studies that rely on fingerprints, CapST processes all facial content, removing the need for supplementary artifacts and enabling effective detection of deep-fake differences.

\begin{algorithm}
\caption{CapST video processing and classification}
\label{capst_algorithm}
\begin{algorithmic}[1]
\STATE \textbf{Input:} A video with $10$ frames, each with $3$ channels and dimension $112 \times 112$
\STATE Initialize an empty tensor $all\_fea$ to store features of all frames

\FOR{each frame $i$ in the 10 frames}
    \STATE \textbf{Feature Extraction:}
    \STATE Extract features $x_i$ using VGG19: $f_i = \text{VGG19}(x_i)$
    
    \STATE \textbf{Capsule Network:}
    \STATE Initialize Capsule Network input $u_i = f_i$
    \STATE Apply a primary convolutional layer to get $u_{i,j}$
    \FOR{each capsule $j$ in the primary capsule layer}
        \STATE Compute the primary capsule output: $u_{i,j} = W_j \cdot u_i + b_j$
        \STATE Apply non-linear squash function: $v_{i,j} = \frac{\|u_{i,j}\|^2}{1 + \|u_{i,j}\|^2} \frac{u_{i,j}}{\|u_{i,j}\|}$
    \ENDFOR
    \STATE Compute agreement for dynamic routing
    \STATE Iterate over dynamic routing steps
    \FOR{iteration $r = 1$ to $R$}
        \STATE Compute coupling coefficients $c_{ij} = \frac{\exp(b_{ij})}{\sum_k \exp(b_{ik})}$
        \STATE Compute capsule output: $s_j = \sum_i c_{ij} v_{i,j}$
        \STATE Apply squash function: $v_j = \frac{\|s_j\|^2}{1 + \|s_j\|^2} \frac{s_j}{\|s_j\|}$
        \STATE Update log prior: $b_{ij} = b_{ij} + v_j \cdot u_{i,j}$
    \ENDFOR
    \STATE Obtain final capsule output: $c_i = v_j$
    
    \STATE Apply average pooling: $p_i = \text{AvgPool}(c_i)$
    \STATE Reshape the output: $r_i = \text{reshape}(p_i, [bs, 256])$
    \STATE Store the reshaped output $r_i$ in $all\_fea$ at index $i$
\ENDFOR

\STATE \textbf{Temporal Attention:}
\STATE Concatenate features of all frames: $F = [r_1, r_2, \ldots, r_{10}]$
\STATE Compute attention weights: $\alpha = \text{TA}(F)$ where $\alpha \in \mathbb{R}^{bs \times 1 \times 10}$
\STATE Normalize attention weights: $\alpha_{sum} = \sum_{t=1}^{10} \alpha_t$
\STATE Weighted feature representation: $F' = \alpha \odot F$
\STATE Compute the final feature representation: $f_{final} = \sum_{t=1}^{10} F'_t$
\STATE Normalize final feature: $f_{final} = \frac{f_{final}}{\alpha_{sum}}$

\STATE \textbf{Classification:}
\STATE Pass the normalized features through a fully connected layer: $y = \text{FC}(f_{final})$
\STATE \textbf{Output:} Classification result $y$ with 5 classes
\end{algorithmic}
\end{algorithm}

\subsection{CapST Algorithm Modules Explanation}
The CapST algorithm consists of several modules.

\noindent \textbf{Feature Extraction} utilizes the first 26 layers of VGG19 to extract features from each frame \( x_i \), producing feature maps \( f_i = \text{VGG19}_{0-26}(x_i) \) with dimensions \( f_i \in \mathbb{R}^{bs \times 256 \times 14 \times 14} \). This 26 layer selection is optimal as it balances performance and computational cost.

\noindent In the \textbf{Capsule Network}, these features are processed through the primary capsule layer and dynamic routing. The primary capsule layer applies a convolution with a weight matrix \( W_j \) and bias \( b_j \), generating activations \( u_{i,j} \). These activations are transformed using a squash function and the Max-Feature-Map (MFM) activation:
\[
v_{i,j} = \frac{\|u_{i,j}\|^2}{1 + \|u_{i,j}\|^2} \frac{u_{i,j}}{\|u_{i,j}\|} \quad \text{and} \quad \text{MFM}(x) = \max(x_1, x_2).
\]
\setlength{\abovedisplayskip}{5pt}
\setlength{\belowdisplayskip}{5pt}
Dynamic routing adjusts the coupling coefficients \( c_{ij} \) with:
\[
c_{ij} = \frac{\exp(b_{ij})}{\sum_k \exp(b_{ik})},
\]
computes the output \( s_j = \sum_i c_{ij} v_{i,j} \), transforms it into the final output \( v_j = \frac{\|s_j\|^2}{1 + \|s_j\|^2} \frac{s_j}{\|s_j\|} \), and updates the log prior \( b_{ij} = b_{ij} + v_j \cdot u_{i,j} \) producing the final output of the capsule \( c_i = v_j \).

\noindent The \textbf{Temporal Attention} module concatenates features from all frames into a tensor \( F = [r_1, r_2, \ldots, r_{10}] \), with \( r_i \) being the reshaped feature vector for each frame. Temporal attention weights \( \alpha \) are computed and normalized as:
\[
\alpha = \text{TA}(F) \quad \text{and} \quad \alpha_{sum} = \sum_{t=1}^{10} \alpha_t.
\]
Weighted features \( F' \) are computed via element-wise multiplication and summing:
\[
F' = \alpha \odot F \quad \text{and} \quad f_{final} = \sum_{t=1}^{10} F'_t,
\]
with normalization:
\[
f_{final} = \frac{f_{final}}{\alpha_{sum}}.
\]

In the \textbf{Classification} step, the feature representation \( f_{final} \) is passed through a fully connected layer:
\[
y = \text{FC}(f_{final}),
\]
mapping the feature vector to class scores, representing the probability that the input video belongs to each of the 5 classes.



\section{EXPERIMENTS AND RESULTS}
\begin{figure}[t!]
  \centering
  \begin{adjustbox}{trim={2cm} {5cm} {2cm} {3cm},clip, width=\linewidth}
    \includegraphics{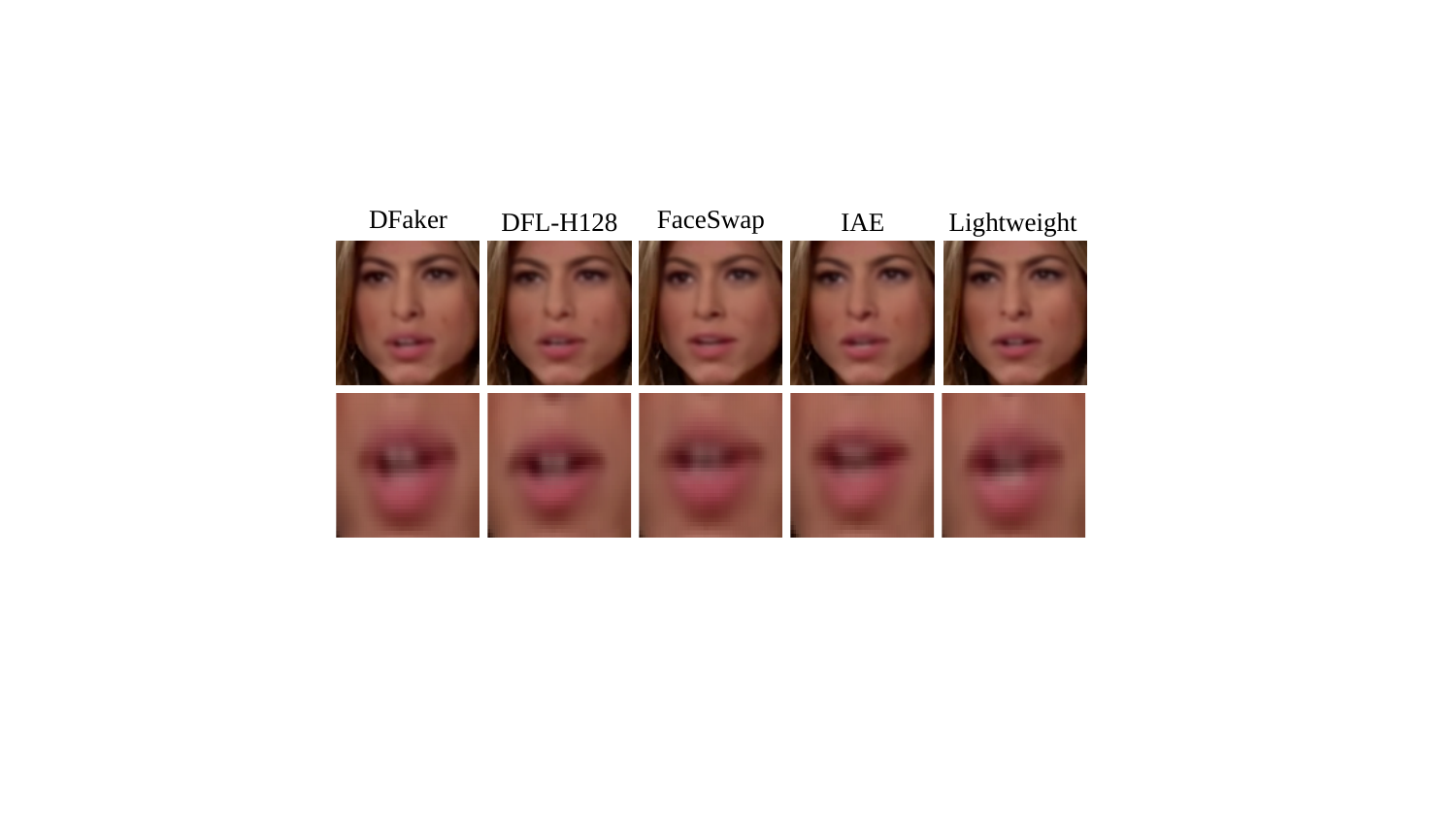}
  \end{adjustbox}
  \caption{DFDM Example: Generated By Different Encoders}
  \Description{ Comparison of deep-fake videos generated using different methods and models, highlighting subtle facial differences and model-specific artifacts. The variations are evident among Deepfakes produced by Faceswap, Lightweight, IAE, Dfaker, and DFL-H128 from the same dataset.}
  \label{dfdm_example}
\end{figure}
\begin{table}
    \centering
    \caption{Summary of Datasets Used in Experiments}
    \begin{tabular}{c|c|c}
        \hline
        \textbf{Dataset} & \textbf{Type of Data} & \textbf{Number of Samples}\\
        \hline
        DFDM & Autoencoder-based Deep-fake videos (CRF10) & 2,150\\
        GANGen-Detection & GAN-based fake images (9 classes) & 18,000 (2,000 per class)\\
        \hline
    \end{tabular}
    \label{datasets_summary}
\end{table}

\subsection{Datasets}

We utilized two key datasets for our experiments: the DeepFakes from Different Models (DFDM) dataset and the GANGen-Detection dataset, selected for their diversity and relevance in evaluating fake detection methods. The details of both datasets are shown in Table~\ref{datasets_summary}.

\noindent The \textbf{DFDM dataset} \cite{jia2022model} contains 6,450 fake videos generated by five models (FaceSwap, Lightweight, IAE, Dfaker, DFL-H128) with varying architectures. These videos, created using the Faceswap tool, are categorized by generation techniques, allowing a detailed evaluation of the model. Figure \ref{dfdm_example} illustrates model-specific artifacts.

\noindent The \textbf{GANGen-Detection dataset} \cite{aigcdataset} includes real and fake samples from GAN architectures such as StyleGAN, StyleGAN2, and ProGAN. This dataset assesses the robustness of detection against high-quality GAN-based manipulations.

\noindent By combining \textbf{DFDM} and \textbf{GANGen-Detection}, we evaluate the robustness of our model with various manipulation techniques. Table \ref{datasets_summary} summarizes the details of the dataset.

\subsection{Experimental Setting}
Before conducting our experiments, we divided the DFDM dataset into training and testing sets following the guidelines provided by \cite{jia2022model}, with a ratio of 70\% for training and 30\% for testing. To extract frames from each video in the dataset, we utilized the OpenFace open-source library \cite{baltrusaitis2018openface}, known for its accuracy and reliability. The extracted frames were saved in PNG format with a dimension of 112 $\times$ 112 pixels. Using a lower dimension, we aimed to reduce the computational resources required to train our model, which is one of the main objectives of this study. To ensure a balanced distribution of faces in frame selection, we employed periodic sampling to select 10 frames from each video for our proposed approach. Our experimental results indicate a clear performance improvement when increasing the number of frames from 1 to 10, while the performance becomes stable beyond 10 frames. Thus, the network takes as input the cropped faces with dimensions of 112 $\times$ 112 $\times$ 3 $\times$ 10. However, for the GANGen-Detection dataset, there was no official split, so we had to split it randomly by 70\% and 30\% as well for training and testing, respectively, while the input dimension was kept as 128 $\times$ 128 as most of the images among all the classes has the dimension of 128 and few of the classes had 256. Similarly, the dimension was not reduced due to the less amount of data that was not creating a hurdle to be proceeded on our provided computational resources, therefore, all the experiments were conducted on an NVIDIA GeForce RTX 3080 Ti with 12 GB of memory for all the models. This choice of hardware allowed us to efficiently evaluate the performance of our model in terms of both accuracy and resource utilization. For training the network, we used the SGD optimizer with a weight decay of 5 $\times$ 10$^{-4}$ and a momentum of 0.9. The learning rate was set to 0.01 throughout the entire training process. A batch size of 10 was used and training was carried out for 300 epochs.
\subsection{Results and Evaluation}
In this study, we evaluate the performance of our model using the class-wise accuracy derived from the confusion matrix. The confusion matrix allows for detailed insights into how well the model performs in different classes by tracking true positives, false positives, true negatives, and false negatives for each category.

\noindent For all the experiments we generated a confusion matrix to compute the accuracy for each class individually. The diagonal elements of the confusion matrix represent the correctly predicted samples for each class, while the off-diagonal elements show the misclassifications. Using this, we calculate the class-wise accuracy as follows.
\[
\text{Accuracy for Class i} = \frac{\text{True Positives for Class i}}{\text{Total Samples for Class i}} \times 100
\]
\begin{itemize}
    \item \textbf{Class-wise Performance:} The individual accuracies across the five classes (FaceSwap, Lightweight, IAE, Dfaker and DFL-H128) are calculated based on the confusion matrix. For example, our model achieves the highest accuracy in the Dfaker class (93.07\%), indicating its robustness in detecting manipulations generated by this model, while it performs moderately in the Lightweight class (53.84\%).
    
    \item \textbf{Insights from Misclassifications:} By examining the confusion matrices, we observe that the misclassification rates are higher between the Lightweight and IAE classes, possibly due to similarities in the deep-fake generation techniques employed by these models. This analysis provides valuable feedback for future work, focusing on reducing false positives for similar models.
\end{itemize}

\noindent It is important to note that the scope of our experiments is limited to a multi-class, fully synthetic image attribution. The CapST model focuses on attributing synthetic images and videos to their respective generation models without performing real-vs-fake classification. This aligns with the composition of the DFDM dataset, which contains 6,450 videos generated using synthetic techniques, providing an ideal setting for controlled evaluations focusing on attribution.
\begin{table*}[!t]
  \centering
  \caption{Comparison of Accuracy (\%) Across State-of-the-Art Models on DFDM Dataset by Class (FaceSwap, LightWeight, IAE, Dfaker, DFL-H128) and Overall Average Accuracy. Each model's performance is reported by class, with the highest accuracy per class bolded for quick reference. An additional row displays the highest accuracy achieved in each class across all models.}
  \label{dfdm_model_comparison}
  \begin{tabular}{lcccccc}
    \hline
    \textbf{Method} & \textbf{FaceSwap} & \textbf{LightWeight} & \textbf{IAE} & \textbf{Dfaker} & \textbf{DFL-H128} & \textbf{Average}\\
    \hline
    MesoInception \cite{afchar2018mesonet} & 6.98 & 2.33 & \textbf{79.07} & 79.07 & 4.65 & 20.93\\    
    R3D \cite{de2020deepfake} & 27.13 & 25.58 & 15.5 & 20.16 & 18.61 & 21.40\\     
    DSP-FWA \cite{li2019exposing} & 17.05 & 7.75 & 43.41 & 40.31 & 8.87 & 23.41\\
    GAN Fingerprint \cite{marra2019gans} & 20.16 & 22.48 & 54.26 & 21.71 & 26.36 & 28.99\\
    DFT-spectrum \cite{durall2019unmasking} & \textbf{99.92} & 3.26 & 0.23 & 27.21 & 48.91 & 35.91\\
    Capsule \cite{8682602} & 32.56 & 42.64 & 69.77 & 73.64 & 58.91 & 55.50\\
    SAM+FA \cite{8803603} & 64.34 & 42.64 & 76.61 & 74.42 & 79.07 & 66.82\\
    ResNet-50 \cite{he2016deep} & 54.84 & 57.36 & 70.54 & 89.92 & 70.54 & 68.02\\
    CBAM \cite{woo2018cbam} & 52.42 & \textbf{63.57} & 69.77 & 84.50 & 74.42 & 68.53\\
    SAM+Ave & 58.87 & 51.16 & 76.74 & 80.62 & 79.07 & 68.84\\
    DMA-STA \cite{jia2022model} & 63.57 & 58.91 & 66.67 & 82.95 & 87.60 & 71.94\\
    \hline
    \textbf{CapST[Ours]} & 77.69 & 53.84 & 60.76 & \textbf{93.07} & \textbf{92.30} & \textbf{75.54}\\
    \hline
    \textbf{High(per class)} & \textbf{99.92} & \textbf{63.57} & \textbf{79.07} & \textbf{93.07} & \textbf{92.30} &\\
    \hline
  \end{tabular}
  \begin{tablenotes}
    \footnotesize
    \item - Each model’s performance across different classes is reported, with CapST achieving the highest average accuracy. Bold values indicate the highest accuracy for each class across all models.
    \item - The "Average" column represents the mean accuracy of each model across all five classes, offering an overall performance metric to compare methods.

  \end{tablenotes}
\end{table*}

\subsubsection{\textbf{Comparison with existing methods on DFDM Dataset}}

We compare our proposed model ``CapST'' with various state-of-the-art methods and present the average accuracy across all five classes, along with the accuracy for individual Deep-fake generation models/classes (e.g., FaceSwap, Lightweight, IAE, etc.) on DFDM dataset high-quality video (CRF10).
Table \ref{dfdm_model_comparison} compares our model with state-of-the-art methods, including attention-based and classification approaches, on high-quality DFDM videos (CRF10). Our model achieves the highest overall accuracy, demonstrating its superior classification performance. Despite some competing models using skip connections and complex architectures, our simple VGG19-based approach performs exceptionally well. Although skip connection-based models may excel in specific classes, they fall short in overall effectiveness. Although ResNet often outperforms VGG19 in large-scale tasks such as ImageNet, VGG19 proves to be robust for our specific dataset and problem. Through experiments, we found that combining the first 26 layers of VGG19 with a Capsule Network yields better performance, as shown in Table \ref{vgg_different_layers_performance} in Section \ref{sec:ablation_study}. In general, our method achieves an average accuracy of 75.54\%, outperforming other methods in various classes, as detailed in Table \ref{dfdm_model_comparison}.


\begin{table}[!t]
    \centering
    \renewcommand{\arraystretch}{1.3}
    \setlength{\tabcolsep}{1pt} 
    \begin{adjustbox}{width=\textwidth}
    \begin{tabular}{c|c|c|c|c|c|c|c|c|c|c}
        \hline
        \textbf{Model} & \textbf{AttnGAN} & \textbf{BEGAN} & \textbf{CramerGAN} & \textbf{InfoMaxGan} & \textbf{MMDGAN} & \textbf{RelGAN} & \textbf{S3GAN} & \textbf{SNGAN} & \textbf{STGAN} & \textbf{Average}\\
        \hline
        EfficientNet & \textbf{100} & \textbf{100} & 99.33 & \textbf{100} & 98.17 & 99.67 & \textbf{100} & 99.83 & \textbf{100} & 99.67\\
        \hline
        DMA-STA & 99.83 & \textbf{100} & \textbf{97.67} & 99.83 & 94.83 & \textbf{100} & 98.50 & 99.83 & \textbf{100} & 98.93\\
        \hline
        CapST (Ours) & \textbf{100} & \textbf{100} & 99.17 & \textbf{100} & \textbf{99.50} & 99.83 & 99.83 & \textbf{99.83} & \textbf{100} & \textbf{99.76}\\
        \hline
    \end{tabular}
    \end{adjustbox}
    \caption{Performance Comparison of CapST with Different Existing Methods on GANGen-Detection Dataset}
    \label{GANGen-Detection_performance_comparison}
\end{table}
\subsubsection{\textbf{Comparison with existing methods on GANGen-Detection Dataset}}
The results in Table \ref{GANGen-Detection_performance_comparison} show a comparison of classification accuracy across videos generated by different GAN models using three methods: DMA-STA, EfficientNet, and CapST (Ours). Our proposed model, CapST, achieves the highest overall performance with an average accuracy of 99.76\%, outperforming both DMA-STA (98.93\%) and EfficientNet (99.67\%). CapST demonstrates superior accuracy in 6 of the 9 GAN models, specifically excelling in AtnGAN, BEGAN, InfoMaxGAN, MMDGAN, SNGAN, and STGAN, where it achieves the best results. CapST also shows remarkable consistency, with its lowest accuracy being 99.17\%, suggesting that it can reliably detect GAN-generated images across a wide range of architectures. In particular, for MMDGAN, CapST achieves 99.50\% accuracy, significantly outperforming DMA-STA (94.83\%) and slightly surpassing EfficientNet (98.17\%). In comparison, while EfficientNet performs well, achieving perfect accuracy in 5 GAN models (AtnGAN, BEGAN, InfoMaxGAN, S3GAN, and STGAN), it is still slightly behind CapST on average. DMA-STA, while competitive, especially with perfect scores in RelGAN and STGAN, falls short in models such as MMDGAN and CramerGAN.

\noindent Overall, these results demonstrate that CapST is not only robust and consistent but also the most effective model in detecting GAN-generated content across a wide variety of architectures, outperforming both DMA-STA and EfficientNet in terms of overall accuracy and consistency.
\begin{figure}
    \centering
    \includegraphics[width=0.5\linewidth]{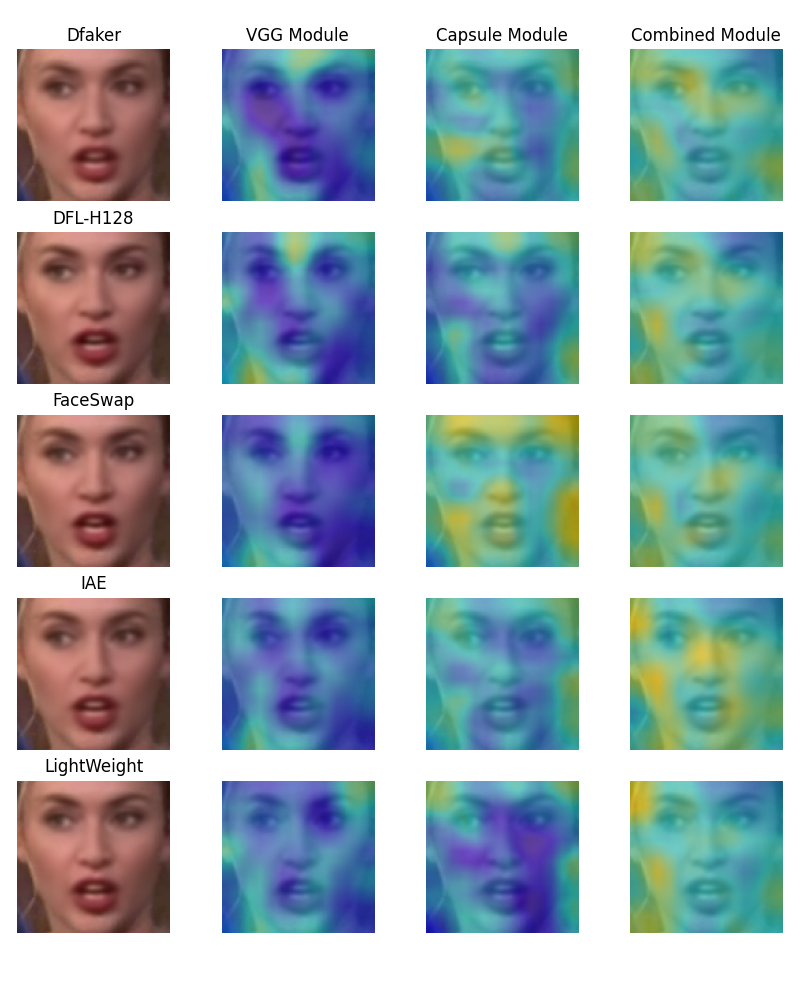}
    \caption{Grad-CAM visualizations of different deep-fake techniques (Dfaker, DFL-H128, FaceSwap, IAE, Lightweight) across three model configurations: VGG Module, Capsule Module, and Combined Module.}
    \Description{Each column represents the activation focus for a given model on manipulated facial regions. The VGG module shows broad facial feature coverage, the Capsule module displays focused attention on localized areas, and the Combined module effectively balances these strengths, providing comprehensive attention to manipulation-prone regions. This highlights the Combined model's superior adaptability and robustness in detecting diverse deep-fake artifacts.}
    \label{gradcam_heatmaps}
\end{figure}


\subsubsection{\textbf{Analysis of Grad-CAM Output Across Different Deep-fake Techniques and Model Configurations}}

 Grad-CAM visualizations in Figure \ref{gradcam_heatmaps} provide insight into the activation patterns of three model configurations: VGG, Capsule, and the combined model. By analyzing where each model focuses on different deep-fake images, we assess the interpretability and effectiveness of each model type in identifying manipulated content.

\noindent \textbf{VGG Module}
The Grad-CAM heatmaps of the VGG module show broad attention throughout the face, with an emphasis on the core facial regions such as the eyes, nose, and mouth. This module provides a generalized feature extraction across facial structures, which is beneficial for identifying large-scale inconsistencies that arise in deep-fake generation. For Dfaker and Lightweight, the VGG module captures some artifacts near the eyes and mouth, areas frequently misaligned in deepfakes. However, its activations are not as concentrated, indicating limitations in pinpointing subtle manipulation artifacts. Overall, the VGG module shows moderate coverage across all techniques, but its generality suggests that it may miss finer details critical for high-fidelity deep-fake detection.

\noindent \textbf{Capsule Module:}
The Capsule module exhibits more precise focus on specific regions, particularly the eyes, mouth, and cheeks, which are often problematic in deep-fake images. This spatial attention enhances its ability to capture localized artifacts that can differentiate authentic from manipulated faces. In the cases of DFL-H128 and FaceSwap, the Capsule model’s heatmaps display focused activations around the eyes and mouth, suggesting it is effective at detecting distortions introduced by these techniques. This refined attention could make the Capsule model more sensitive to intricate manipulation details. However, the Capsule module alone shows a somewhat narrow focus, which might limit its performance on techniques that require a broader spatial understanding, as seen in its response to the IAE method.

\noindent \textbf{Combined Model (VGG + Capsule):}
The Combined model, representing a synthesis of the VGG and Capsule modules, balances a wide coverage of features with focused spatial attention. This configuration displays the most comprehensive and consistent activation patterns across all deep-fake types. For Dfaker and Lightweight, the combined model shows significant attention to the mouth, eyes, and chin regions, indicating that it effectively identifies areas commonly altered in deep-fake techniques. The heat maps of the combined model reveal a robust ability to capture both localized and broad artifacts, making it adaptable to the diverse characteristics of different deep-fake methods. This configuration leverages both VGG’s broad feature extraction and Capsule’s fine-grained attention, leading to higher precision and interpretability in deep-fake detection.

\noindent \textbf{Comparative Analysis Across Deep-fake Techniques:}
\begin{itemize}
    \item \textbf{Dfaker and DFL-H128:} Capsule and paired models show strong activations around the mouth and eyes, which are frequently distorted in these techniques. The combined model performs particularly well, as it utilizes both high-level structure and detailed spatial consistency to detect subtle artifacts.
    \item \textbf{FaceSwap:} The Capsule model is highly effective in focusing on regions affected by the FaceSwap method, such as the nose and cheeks. However, the combined model provides a more comprehensive view, capturing artifacts across the face, and enhancing detection performance for this technique.
    \item \textbf{IAE and Lightweight:} These methods present a challenge due to improved blending and reduced artifacts. The broad focus of the VGG model is somewhat effective for light weight, but the combined model outperforms it by maintaining a precise focus on key facial areas, indicating adaptability to advanced deep-fake techniques with fewer visible artifacts.
\end{itemize}
\textbf{Implications for Deep-fake Detection:}
The Combined model’s consistent and well-distributed focus on the combined model in facial features demonstrates its superiority in detecting manipulations across various deep-fake methods. Its ability to take advantage of both VGG's general feature extraction and Capsule’s spatial precision makes it highly effective for general-purpose deep-fake detection. The Grad-CAM outputs confirm that the combined model is more interpretable and reliable for real-world applications where different deep-fake techniques may be encountered. By effectively balancing global and local feature extraction, the Combined model maximizes both sensitivity and specificity, crucial attributes for robust deep-fake detection.

\noindent \textbf{Conclusion:}
The Grad-CAM analysis supports the conclusion that combining VGG and Capsule modules results in a more adaptable and effective deep-fake detection model. The heat maps of the combined model highlight its strengths in identifying the manipulated regions consistently with multiple techniques, demonstrating it as the most balanced approach. This analysis emphasizes the advantage of the combined model design, which can lead to both high detection accuracy and improved model interpretability.

\subsubsection{\textbf{Limitations and Observations:}}

Although the CapST model demonstrates robust performance across multiple classes and datasets, certain limitations and failure cases were identified.

\begin{enumerate}
    \item \textbf{Edge Cases:} The model struggles to distinguish between lightweight and IAE classes, as evident from the higher misclassification rates in the confusion matrix. This is likely due to similarities in the generation techniques employed by these models, resulting in overlapping artifacts that are challenging to separate.
    
    \item \textbf{Higher GFLOPs of VGG:} The CapST model leverages the VGG architecture for feature extraction, which contributes to its higher GFLOP count (29.54 GFLOPs) compared to other models such as DMA-STA (1.35 GFLOPs) and EfficientNet (3.04 GFLOPs). One key reason for this increase is the absence of skip connections in the VGG architecture, unlike architectures such as ResNet. Skip connections enable direct gradient flow, reducing redundancy and improving computational efficiency, whereas VGG processes each layer sequentially. This design choice, while increasing GFLOPs, allows for a more thorough feature extraction process that complements the Capsule Network's ability to capture spatial and temporal attention. Despite the higher GFLOPs, this trade-off results in superior detection accuracy while maintaining a compact parameter count (3.27M), making it a competitive model for deepfake attribution tasks.
\end{enumerate}

\noindent These limitations highlight areas for further research and improvement, particularly in enhancing the model's ability to detect subtle or blended artifacts and improving differentiation between similar generation techniques.

\section{Ablation Study} \label{sec:ablation_study}
The CapST module demonstrates computational efficiency and high performance through four core phases. First, frames are extracted from each video and processed using VGG19 for feature extraction. VGG19 was selected over ResNet due to its better compatibility with the Capsule network, despite ResNet’s parameter efficiency. To optimize the parameters and obtain better performance, we only use the first three max-pooling layers of VGG19, as shown in Table \ref{vgg_different_layers_performance}. These features were then enhanced through the Capsule module, which strengthens spatial characteristics. Finally, video-level fusion facilitates accurate video classification, distinguishing CapST from previous approaches by reducing image dimensions, batch size, and training time. These optimizations improve the classification of fake content in FPGA-based hardware. We conducted several experiments to ablate our study to further investigate the effectiveness of our proposed solution. 
\begin{table}[!t]
  \centering
  \begin{threeparttable}
    \caption{Performance Comparison of Our Model and DMA-STA on Reproduced Results Using Consistent Experimental Settings on DFDM Dataset. The best score is marked in bold.}
    \label{tab:reproduced_results}
    \begin{tabularx}{\textwidth}{l*{7}{X}}
      \toprule
      \textbf{Method} & \textbf{lr} & \textbf{FS} & \textbf{LW} & \textbf{IAE} & \textbf{Dfaker} & \textbf{DFL} & \textbf{Avg}\\
      \midrule
      DMA-STA$^*$ \cite{jia2022model} & $1 \times 10^{-2}$ & 53.84 & 48.46 & 61.53 & 76.92 & 63.84 & 60.92\\
      CapST: Orig\_Caps & $1 \times 10^{-4}$ & 70.00 & 46.89 & 73.07 & 90.00 & 83.84 & 72.76\\
      CapST: Orig\_Caps & $1 \times 10^{-2}$ & 72.30 & 38.46 & \textbf{76.92} & 87.69 & 91.53 & 73.38\\
      CapST: MFM\_5$\times$5 & $1 \times 10^{-2}$ & 72.30 & 42.30 & 73.84 & 90.76 & 91.53 & 74.15\\
      \textbf{CapST: MFM\_3$\times$3} & $1 \times 10^{-2}$ & \textbf{77.69} & \textbf{53.84} & 60.76 & \textbf{93.07} & \textbf{92.30} & \textbf{75.54}\\
      \bottomrule
    \end{tabularx}
    \begin{tablenotes}
      \footnotesize
      \item[1] DMA-STA$^*$ reproduced results of the baseline method on our hardware.
      \item[2] lr, FS, LW, and DFL represent Learning Rate, FaceSwap, Lightweight, and DFL-H128, respectively.
      \item[3] 5 $\times$ 5 and 3 $\times$ 3 are kernel sizes.
      \item[4] MFM (Max-Feature-Map) Activation Function.
    \end{tablenotes}
  \end{threeparttable}
\end{table}
\subsection{Comparing Our Results to DMA-STA Reproduced Results} 
While the initial performance of DMA-STA \cite{jia2022model} was promising, our model demonstrated significant superiority after applying our custom settings. However, we also further verify the performance of the existing baseline by running the model on our own data under the same setting that we applied to our model. The reason for conducting such an experiment was to investigate the performance of the baseline under the same hardware and hyperparameters due to the limited resources available to us. As our machine's limited GPU memory prevented us from using their original settings therefore, to optimize model efficiency and resource usage, we adjusted the original settings that involved image dimensions from the video dataset and batch size as shown in Table \ref{tab:reproduced_results}. The original DMA-STA experiments employed Titan XP GPUs \cite{deepfake_model_attribution_issue2}. However, we had access to only a single GPU with 12 gigabytes of memory. To ensure a fair comparison, we adapted the same hyper-parameters and aligned only the image dimension and batch size to reproduce DMA-STA's results on our machine. Specifically, we reduced image dimensions from $300 \times 300$ to $112 \times 112$ and the batch size from 20 to 10 while keeping other settings unchanged.
\begin{table}[!t]
  \centering
  \begin{threeparttable}
    \caption{Parameters and Training Time Comparison of DMA-STA and CapST Models on RTX-3080ti}
    \label{computational_complexity}
    \setlength{\tabcolsep}{4pt} 
    \renewcommand{\arraystretch}{1.1} 
    \begin{tabular}{llccc}
      \toprule
      \textbf{Index} & \textbf{Model Layers} & \textbf{Params (M)} & \textbf{FLOPs (G)} & \textbf{Training Time (Min)}\\
      \midrule
      \multicolumn{5}{l}{\textbf{DMA-STA \cite{jia2022model}}}\\
      0 & Conv2D           & 9.40   & -     & -     \\
      1 & BatchNorm        & 0.128  & -     & -     \\
      2 & Layer 1-4        & 23.500 & -     & $0.5025 \times 300$\\
      3 & TA-FC0           & 0.05   & -     & -     \\
      4 & TA-FC1           & 0.05   & -     & -     \\
      5 & TA-FC            & 10.245 & -     & -     \\
      \midrule
      \multicolumn{2}{l}{\textbf{Total}} & 23.52 & 10.87 & 150.75\\
      \midrule
      \multicolumn{5}{l}{\textbf{CapST [Ours]}}\\
      0 & VGG              & 2.328  & -     & -     \\
      1 & CapsNet$\times3$ & 0.942  & -     & -     \\
      2 & TA-FC0           & 0.05   & -     & $0.373 \times 300$\\
      3 & TA-FC1           & 0.05   & -     & -     \\
      4 & TA-FC            & 2.500  & -     & -     \\
      \midrule
      \multicolumn{2}{l}{\textbf{Total}} & 3.27  & 29.54 & 111.9\\
      \bottomrule
    \end{tabular}
    \begin{tablenotes}
      \footnotesize
      \item[1] Training Time$^*$ is calculated for 300 epochs in each experiment.
      \item[2] TA-FC: Temporal-Attention Fully Connected Layers.
    \end{tablenotes}
  \end{threeparttable}
\end{table}

\subsection{Comparison with DMA-STA in terms of Computational Complexity} 

The DMA-STA network, using ResNet for feature extraction and Spatial-Temporal attention modules, achieves high-level representations but results in a substantial parameter count of 23.520 million. In contrast, our CapST model employs a simpler VGG-based feature extractor and a capsule network, reducing the parameters to just 3.27 million, eight times smaller than DMA-STA, while providing comparable or superior performance. Although CapST's computational complexity increases to 29.54 GFLOPs due to CapsNet layers, it compensates with faster training, completing in 111.9 minutes versus DMA-STA's 150.75 minutes, a 25.8\% reduction. Table \ref{computational_complexity} summarizes these findings.

\noindent Table \ref{GANGen-Detection:model_params_gflops} compares the CapST model with DMA-STA and EfficientNet in terms of parameters and GFLOPs. CapST has the smallest parameter count (3.86M), outperforming EfficientNet (4.02M) and DMA-STA (23.53M). With 3.04 GFLOPs, CapST also has a much lower computational cost than EfficientNet (135.57 GFLOPs) while remaining practical compared to DMA-STA, which, despite its low GFLOPs (1.35), has a significantly larger parameter size. CapST’s low parameters and balanced GFLOPs make it efficient for real-world applications, achieving competitive performance with fewer resources than EfficientNet and DMA-STA.

\begin{table}[!t]
    \centering
    \caption{Comparison of Model Parameters and GFLOPs on GANGen-Detection Dataset}
    \begin{tabular}{c|c|c}
        \hline
        \textbf{Model} & \textbf{Params (M)} & \textbf{FLOPs(G)}\\
        \hline
        EfficientNet & 4.02 & 135.57\\
        \hline
        DMA-STA & 23.53 & \textbf{1.35}\\
        \hline
        CapST (Ours) & \textbf{3.86} & 3.04\\
        \hline
    \end{tabular}
    \label{GANGen-Detection:model_params_gflops}
\end{table}

\subsection{Comparison using different backbones}
We also evaluate our proposed method on different backbones for comparative analysis of various, highlighting CapST's advantages in both computational efficiency and classification accuracy. Among the models evaluated, EfficientNet and MobileNet-V2 stand out for their low FLOPs (1.14G and 1.63G, respectively), with EfficientNet achieving a competitive accuracy of 71.69\% and the shortest training time of 90 minutes. However, despite its efficiency, EfficientNet falls short of CapST's overall performance. CapST, while requiring a slightly longer training time of 112 minutes, maintains a balanced profile with 3.27 million parameters and 29.54 FLOPs. Most notably, CapST exceeds all other models in accuracy, achieving an impressive 75.54\%, which is 3.85\% higher than EfficientNet, the next best-performing model. This performance improvement demonstrates CapST’s robustness in classification accuracy while still keeping computational demands at reasonable levels, as shown in Table \ref{modified_backbones}. The comparison underscores CapST's strength in achieving a high accuracy-to-efficiency ratio, making it a well-optimized solution for applications requiring both effective and efficient backbone architectures.

\begin{table}[!t]
  \centering
  \caption{Comparison by Changing Different Backbones(Acc\%). Best Results are marked in bold.}
  \label{modified_backbones}
  \begin{threeparttable}
    \begin{tabular}{lcccc}
      \toprule
      \textbf{Model} & \textbf{Training Time (min)} & \textbf{Params (M)} & \textbf{FLOPs (G)} & \textbf{Avg: Acc (\%)}\\
      \midrule
      XceptionNet \cite{chollet2017xception} & 120 & 20.81 & 11.61 & 38.61\\
      ResNet-152 \cite{he2016deep} & 268 & 57.03 & 29.55 & 60.46\\
      MobileNet-V2 \cite{sandler2018mobilenetv2} & 118 & \textbf{2.23} & 1.63 & 21.69\\
      InceptionNet \cite{szegedy2015going} & 126 & 6.27 & 12.9 & 42.76\\
      EfficientNet \cite{tan2019efficientnet} & \textbf{90} & 4.01 & \textbf{1.14} & 71.69\\
      \midrule
      \textbf{CapST[Ours]} & 112 & \textbf{3.27} & 29.54 & \textbf{75.54}\\
      \bottomrule
    \end{tabular}
    \begin{tablenotes}
      \small
      \item[1] Training Time$^*$ is presented in Minutes.
      \item[2] Parameters$^*$ are presented in Millions.
    \end{tablenotes}
  \end{threeparttable}
  
\end{table}
\begin{table}[!t]
  \centering
  \caption{Performance comparison across different CapST VGG models with varying number of layers.}
  \label{vgg_different_layers_performance}
  \resizebox{\textwidth}{!}{ 
  \begin{tabular}{l|c|c|c|c|c|c|c|c}
    \hline
    \textbf{Model} & \textbf{FS} & \textbf{LW} & \textbf{IAE} & \textbf{Dfaker} & \textbf{DFL} & \textbf{Overall} & \textbf{Params (M)} & \textbf{GFLOPs}\\
    \hline
    CapST\_18 Layers & 20.77 & 29.23 & 61.54 & 85.38 & 79.23 & 65.23 & 1.86 & \textbf{24.42} \\
    CapST\_22 Layers & 33.85 & \textbf{56.92} & \textbf{76.15} & 88.46 & 73.85 & 65.85 & 2.45 & 29.05 \\
    CapST\_30 Layers & 43.08 & 56.15 & 70.00 & 89.23 & 80.77 & 67.85 & 8.6 & 39.01 \\
    CapST\_34 Layers & 67.69 & 38.44 & 58.46 & 82.31 & \textbf{93.85} & 68.15 & 10.96 & 43.64 \\
    \textbf{CapST\_26 Layers} & \textbf{77.69} & 53.84 & 60.76 & \textbf{93.07} & 92.30 & \textbf{75.54} & 3.27 & 29.54 \\
    \hline
  \end{tabular}
  }
\end{table}
\begin{table*}[!ht]
  \centering
  \caption{Performance comparison of CapST model using different number of frames.}
  \label{different_frames_performance}
  \resizebox{\textwidth}{!}{ 
  \begin{tabular}{l|c|c|c|c|c|c|c|c}
    \hline
    \textbf{Model} & \textbf{FS} & \textbf{LW} & \textbf{IAE} & \textbf{Dfaker} & \textbf{DFL} & \textbf{Overall} & \textbf{Params (M)} & \textbf{GFLOPs}\\
    \hline
    CapST\_6-Frames & 44.83 & \textbf{63.72} & \textbf{80.77} & 61.54 & \textbf{92.97} & 65.85 & 3.27 & \textbf{18.18} \\
    CapST\_8-Frames & 54.62 & 53.85 & 55.38 & 84.62 & 85.38 & 66.77 & 3.27 & 24.24 \\
    CapST\_12-Frames & 54.62 & 40.77 & 70.77 & 83.08 & 83.85 & 66.62 & 3.27 & 36.36 \\
    CapST\_14-Frames & 46.15 & 44.62 & 64.62 & 86.92 & 86.15 & 65.69 & 3.27 & 42.42 \\
    \textbf{CapST\_10-Frames} & \textbf{77.69} & 53.84 & 60.76 & \textbf{93.07} & 92.30 & \textbf{75.54} & 3.27 & 29.54 \\
    \hline
  \end{tabular}
  }
\end{table*}

\subsection{Comparison using different VGG Layers and Number of Frames}
Further insights are provided in Table \ref{vgg_different_layers_performance}, where variations in the VGG layers were explored to assess the capacity of the model to extract features and optimize performance. The best performance was achieved by CapST with 26 layers, reaching an overall accuracy of 75.54\%, while effectively balancing parameter counts and GFLOPs to remain within practical computational limits. This configuration highlights optimal depth to maximize accuracy without excessive computational overhead. Similarly, as detailed in Table \ref{different_frames_performance}, the CapST model was evaluated with varying frame counts to gauge its performance stability in temporal input. The model demonstrated stable accuracy levels across these configurations, with the highest accuracy of 75.54\% achieved using 6 frames and 29.54 GFLOPs. This result underscores the model's adaptability and efficiency, maintaining robust performance across different input frame configurations and thereby validating its suitability for resource-constrained environments.

\section{CONCLUSION}
Our study presents CapST, an advanced and efficient technique for attribution of deep-fake models, specifically designed to address the need for a resource-effective and accurate detection of deep-fake content origins. By selectively incorporating the first three layers of VGG19 for feature extraction and combining them with Capsule Networks (CapsNet) to capture detailed hierarchical relationships, CapST demonstrates strong discriminatory power in identifying distinct deep-fake attributes. The integration of video-level fusion with a temporal attention mechanism further leverages temporal dependencies, significantly enhancing the model's predictive accuracy. Our empirical validation on a robust deep-fake dataset highlights CapST’s superiority over baseline models, achieving improved accuracy while maintaining low computational costs, critical factors for practical application, and scalability in real world scenarios. As the proliferation of synthetic media continues to pose serious security challenges, CapST stands out as a timely and innovative solution, offering enhanced capabilities in both detection and prevention strategies. This research establishes a solid foundation for future work, with the potential to expand CapST’s adaptability to emerging deep-fake generation models, thus fortifying defenses against evolving threats in the synthetic media landscape.

\begin{acks}
This work was supported in part by the National Science and Technology Council under grant nos. \texorpdfstring{112-2223-E-001-001, 111-2221-E-001-013-MY3, 111-2923-E-002-014-MY3, 112-2927-I-001-508, 113-2221-E-004-006-MY2, 113-2622-E-004-001, 113-2221-E-004-001-MY3, 112-2634-F-002-005, 113-2634-F-002-008, 113-2923-E-A49-003-MY2}{NSC Grants} and Academia Sinica under grant no. \texorpdfstring{AS-IA-111-M01}{Academia Sinica Grant}.
\end{acks}

\printbibliography
\end{document}